\documentclass[10pt, a4paper, logo]{googledeepmind} 

\usepackage{amsmath,amsfonts,bm}
\usepackage{graphicx}
\usepackage{booktabs}
\usepackage{hyperref}
\usepackage{multirow}
\usepackage[numbers]{natbib}
\usepackage{caption}
\usepackage[capitalize,noabbrev]{cleveref}
\usepackage[normalem]{ulem}
\usepackage{tabularx}
\usepackage{siunitx}
\usepackage{array}
\usepackage{ragged2e}

\def\eqref#1{equation~\ref{#1}}

% ...

% ...

% ...

\title{QuarkMed Medical Foundation Model Technical Report}

\author{Ao Li}
\author{Bin Yan}
\author{Bingfeng Cai}
\author{Chenxi Li}
\author{Cunzhong Zhao}
\author{Fugen Yao}
\author{Gaoqiang Liu}
\author{Guanjun Jiang}
\author{Jian Xu}
\author{Liang Dong}
\author{Liansheng Sun}
\author{Rongshen Zhang}
\author{Xiaolei Gui}
\author{Xin Liu}
\author{Xin Shang}
\author{Yao Wu}
\author{Yu Cao}
\author{Zhenxin Ma}
\author{Zhuang Jia}

\affil{Quark Medical Team, Alibaba Group}

\begin{abstract}
Recent advancements in large language models have significantly accelerated their adoption in healthcare applications, including AI-powered medical consultations, diagnostic report assistance, and medical search tools. However, medical tasks often demand highly specialized knowledge, professional accuracy, and customization capabilities, necessitating a robust and reliable foundation model. QuarkMed addresses these needs by leveraging curated medical data processing, medical-content Retrieval-Augmented Generation (RAG), and a large-scale, verifiable reinforcement learning pipeline to develop a high-performance medical foundation model. The model achieved 70\% accuracy on the Chinese Medical Licensing Examination, demonstrating strong generalization across diverse medical benchmarks. QuarkMed offers a powerful yet versatile personal medical AI solution, already serving over millions of users at \href{https://ai.quark.cn}{https://ai.quark.cn}.
\end{abstract}

\begin{document}

\maketitle

\section{Introduction}
\label{sec:intro}

The advent of large language models (LLMs) has marked a pivotal moment in artificial intelligence, demonstrating remarkable capabilities in understanding and generating human-like text across a multitude of domains. This progress has catalyzed significant interest in their application to specialized fields, particularly medicine, where they hold the potential to revolutionize medical information retrieval, enhance early diagnostic accuracy, and support personalized health care requirements.

However, the medical domain presents unique and formidable challenges \cite{xie2025medical}. Unlike general-domain text, medical language is characterized by a highly specialized vocabulary, complex clinical concepts, and a nuanced syntax that is often ambiguous and context-dependent. As a result, general-purpose LLMs, which are typically fine-tuned for broad, non-medical corpora, often lack the deep, specialized knowledge required for high-stakes medical applications \cite{azizi2022medpalm}. This knowledge gap can lead to unsatisfactory, and at times unsafe, performance when these models are directly applied to medical tasks.

Recognizing these limitations, the research community has shifted towards developing domain-specific foundation models for medicine. This endeavor began with the adaptation of Transformer-based architectures like BERT (Bidirectional Encoder Representations from Transformers). Early pioneering work led to the creation of models such as BioBERT \cite{lee2019biobert}, which was pre-trained on large-scale biomedical literature, and ClinicalBERT \cite{huang2019clinicalbert}, which was trained on unstructured clinical notes from electronic health records (EHRs). These models demonstrated that domain-specific pre-training significantly improves performance on various biomedical text mining tasks. Following this trend, models like BEHRT were developed to specifically model structured EHR data for predicting clinical events \cite{li2020behrt}.

The success of these earlier models paved the way for the development of generative models tailored for medicine. BioGPT, for instance, was a generative pre-trained transformer that excelled at creating fluent biomedical text and improving performance on downstream tasks \cite{luo2022biogpt}. As model scaling became a key driver of performance, the field saw the emergence of significantly larger and more powerful medical LLMs. Models like GatorTron, with billions of parameters trained on massive clinical text datasets, demonstrated the benefits of scale in capturing the long-range dependencies and intricate relationships within clinical narratives \cite{yang2022gatortron}.

More recently, the landscape has been defined by even larger and more sophisticated models that integrate extensive medical knowledge with robust instruction-following capabilities. Med-PaLM and its successor were among the first to approach expert-level performance on medical licensing examination-style questions, leveraging a combination of improved base models, medical domain fine-tuning, and advanced prompting strategies \cite{azizi2022medpalm, singhal2023towards}. Concurrently, the open-source community has produced a variety of powerful medical LLMs. Models like PMC-LLaMA \cite{wu2023pmcllama}, MEDITRON-70B \cite{chen2023meditron}, BioMedLM \cite{peng2024biomedlm}, and BioMistral \cite{labrak2024biomistral} have been developed by pre-training on vast corpora of biomedical literature and clinical data, showing performance competitive with proprietary models. This proliferation of models has been accompanied by the creation of more comprehensive and challenging benchmarks, such as MedExQA \cite{desai2024medexqa} and MedS-Bench \cite{wu2025medsbench}, which evaluate LLMs on more complex, long-form question answering and a wider array of clinical tasks.

Beyond supervised learning, Reinforcement Learning (RL) has emerged as a powerful paradigm for optimizing sequential decision-making, making it a promising approach for healthcare applications such as developing dynamic treatment regimes \cite{pubmed2025reinforcement}. Concurrently, Reinforcement Learning from Human Feedback (RLHF) is being explored to incorporate the nuanced expertise of clinicians into the training loop. This allows models to learn from expert guidance, helping to create AI systems that are better aligned with human values and clinical best practices, ultimately enhancing the safety and reliability of the next generation of medical foundation models. However, applying RL in medicine is fraught with challenges, including the need for vast amounts of high-quality data, the difficulty in defining accurate reward functions, and ensuring the interpretability and safety of the models \cite{researchgate2024data}.

To address these limitations, recent advancements have focused on making RL more reliable and verifiable. One such advancement is Reinforcement Learning with Verifiable Rewards (RLVR), which trains models using objective feedback where the correctness of an output can be unambiguously determined, thereby mitigating the risk of "reward hacking" \cite{arxiv2025crossing}. While RLVR has been effective in fields like mathematical reasoning , research is underway to adapt it to the complexities of medicine, where simple binary verification is often insufficient. 

While these advancements are significant, there remains a critical need for medical foundation models that are not only knowledgeable and accurate but also highly reliable and customizable for real-world medical applications. This report introduces QuarkMed, a medical foundation model designed to meet these demands. By leveraging meticulously curated medical data, advanced Retrieval-Augmented Generation (RAG) for verifiable and up-to-date information, and a multi-stage training process including large-scale reinforcement learning, QuarkMed aims to provide a robust and versatile solution for a new generation of AI-powered healthcare tools.

We summarize the main contributions of this work:
\begin{itemize}
  \item \textbf{Comprehensive Medical Data Pipeline:} A multi-layer curation and quality enhancement system (materials, structured knowledge, clinical records) with expert-guided coverage tracking and knowledge synthesis.
  \item \textbf{Ability- and Problem-Driven Instruction Tuning:} A multi-task IFT/SFT curriculum with automated ratio optimization and robustness-oriented adversarial augmentation.
  \item \textbf{Dual-Stage Reinforcement Learning:} A reasoning-focused verifiable reward phase followed by general alignment using multi-dimensional reward (honesty, helpfulness, consistency, compliance) and GRPO in medical domain.
  \item \textbf{Integrated Medical RAG:} Dense retrieval over authority-ranked corpora yielding large factuality and hallucination reductions with citation support.
  \item \textbf{State-of-the-Art Performance:} Strong results across public and internal medical exams (e.g., 70\% Chinese Medical Licensing style accuracy) at a competitive 32B scale.
\end{itemize}

\section{Data}
\label{sec:data}

On top of a general-purpose large language model, challenges remain in the model's parametric knowledge in specific medical domains, which can undermine the performance of downstream tasks. To bridge this gap, we employ a large-scale data processing pipeline to systematically prepare and ingest this knowledge into our base model. Three main types of medical-related data were used during different stages of model training: medical materials, medical knowledge, and medical records. This data contributes to the timely medical knowledge and detailed clinical knowledge for the model, complementing search-based augmented methods.

\subsection{Medical Materials}
To enhance our model's medical expertise, we have collaborated with an internal team of medical experts to build a large-scale, high-quality, and diverse dataset of medical materials. We also employ data synthesis to supplement knowledge points in critical areas.

\paragraph{Data Coverage and Scope}
Through various means such as web crawling and procurement, we have collected a wide range of data including textbooks, clinical guidelines, consensus statements, academic literature, drug inserts, medical encyclopedias, and clinical pathways. This effort has established a comprehensive medical materials library that provides approximately 1T tokens for model training. Based on a framework manually curated by medical experts, we implemented a fine-grained knowledge point coverage detection system. Drawing inspiration from Bloom's Taxonomy, we classify knowledge points into factual, conceptual, and procedural categories. Each category is evaluated using test sets built from Quark search query-cot and our internal medical knowledge graph. Through an iterative process of evaluation and supplementation, our final library achieves over 90\% coverage for factual knowledge, 84\% for conceptual knowledge, and 75\% for qcot coverage. This progressive coverage, from foundational data to complex reasoning data, aligns with the different stages of model training.

\paragraph{Data Quality Enhancement}
A significant portion of the materials exists in image format. Initially, we used OCR and layout analysis models to extract text. To further improve extraction from images with complex layouts or backgrounds, such as those in popular science materials, we trained a fine-grained content structuring model based on qwen2.5 vl. This advanced approach improved the quality of the pre-training corpus by over 30\% compared to the original OCR methods, achieving an average data usability rate of over 90\%, with the rate approaching its upper limit for well-structured images like those in textbooks.

\paragraph{Authoritativeness and Verification}
To ensure the accuracy and reliability of the data, we established an authoritativeness labeling strategy based on the principles of evidence-based medicine. Based on factors such as the material's type, source, and impact factor, we classify data into five authority levels (A-E). This classification is used for filtering data in different stages, such as training and RAG. Within our library, high-authority data accounts for over 40\% of clinical guidelines, 26\% of literature, and 5\% of books.

\paragraph{Knowledge Synthesis for Conceptual Gaps}
The content of original source materials may not adequately address certain high-level conceptual knowledge points. To address this, we employ data synthesis in key medical subdomains—such as diseases, symptoms, drugs, procedures, and tests—by systematically creating knowledge points for a nearly exhaustive set of entities across important relationships defined in our terminology sets. For instance, using data from drug regulatory agencies, we merge the inserts for the same generic drug from various manufacturers, combine this with pharmacological information from encyclopedias and textbooks, and synthesize a comprehensive insert for each generic drug name to be used for general knowledge enhancement.

\subsection{Medical Knowledge}

In the field of healthcare, it is crucial for language models to incorporate a certain level of medical background knowledge for the following reasons:
\begin{itemize}
    \item \textbf{Improving Accuracy and Reducing Hallucinations:} A lack of professional expertise may lead to semantic confusion, resulting in inaccurate or misleading outputs. Incorporating domain-specific knowledge helps ensure that model predictions are grounded in established medical facts.
    \item \textbf{Enhancing Reasoning Capabilities:} Knowledge integration enables the model to perform more sophisticated reasoning tasks, such as inferring potential diseases from a set of symptoms or recommending appropriate diagnostic procedures.
    \item \textbf{Compensating for Gaps in Pre-training Corpora:} Certain critical information—such as data on rare diseases, newly developed drugs, or the latest clinical guidelines—may not be adequately represented in general-domain pre-training datasets. Supplementing with structured medical knowledge helps bridge these gaps and improves the model’s relevance and applicability in real-world clinical settings.
\end{itemize}

\paragraph{Medical Data and Unstructured Data Processing}
The knowledge integrated into the training of large medical language models comes from multiple structured and unstructured sources, categorized based on content type and usage scenario. The classification and approximate scale of each category are summarized in \cref{tab:medical_data_sources}. For unstructured data, it is used in stages such as continued pre-training, instruction fine-tuning, supervised fine-tuning, and reinforcement learning, based on method-specific data selection processes.

\begin{table}[h!]
\centering
\caption{Classification and Scale of Medical Data Sources}
\label{tab:medical_data_sources}
\renewcommand{\arraystretch}{1.2}
\begin{tabular}{lp{6cm}p{4cm}}
\toprule
\textbf{Main Category} & \textbf{Subcategories} & \textbf{Scale} \\
\midrule
Web-based Resources & Q\&A platforms, articles, encyclopedic entries & Tens of millions \\
Professional Materials & Clinical guidelines, publications, drug inserts, medical standards, medical exams & Millions \\
Knowledge Bases & Standard medical terminology sets, medical ontologies, dictionaries & Tens of millions \\
Medical Scenario Data & Online consultation dialogues, patient case records & Tens of millions \\
Other Supporting Data & Legal regulations, medical AI-related databases, clinical trial databases, & Millions \\
& doctor-patient communication data & \\
\bottomrule
\end{tabular}
\end{table}

\paragraph{Knowledge Transformation for Structured Data}
Since LLMs cannot directly use structured data, we transform it into natural language data using a knowledge transformation technique. The process of knowledge selection and construction follows a set of key criteria to ensure its effectiveness in supporting medical language understanding and reasoning. The selected knowledge must meet the following standards:
\begin{itemize}
    \item \textbf{Importance:} The knowledge covers core medical concepts and relationships that are clinically significant.
    \item \textbf{Completeness:} It provides comprehensive coverage of relevant domains and avoids critical omissions.
    \item \textbf{Accuracy and Clarity:} Information is precise, well-defined, and free from ambiguity.
    \item \textbf{Generalization Ability:} The knowledge supports not only direct retrieval but also logical inference and reasoning over unseen or complex scenarios.
\end{itemize}

To align structured knowledge—such as Subject-Predicate-Object (SPO) triples from knowledge graphs—with the model’s native capabilities, knowledge translation techniques are employed. These techniques convert structured SPO data into unstructured natural language sentences, ensuring compatibility with the model’s processing pipeline and enabling more effective integration and learning. This process includes the following key steps:
\begin{itemize}
    \item \textbf{Training the Translation Model:} A translation model is trained to map structured SPO triples into fluent, semantically equivalent natural language sentences. This involves constructing a parallel corpus of SPO triples and their corresponding textual descriptions and training a sequence-to-sequence model to learn the mapping.
    \item \textbf{Extracting Triples from Text and Performing Back-Translation:} To enrich the knowledge corpus and verify its consistency, the system performs triple extraction from unstructured medical texts and back-translation, where the extracted triples are re-translated into natural language to assess their coherence and correctness.
    \item \textbf{Quality Filtering:} A robust quality filtering mechanism is essential to maintain high data standards. This includes semantic consistency checks, grammatical and fluency evaluation, and domain relevance filtering.
\end{itemize}

\paragraph{Evaluation of Knowledge Integration}
To evaluate the effectiveness of this data, we use single-shot methods to probe the parametric knowledge in the model.

\begin{itemize}
    \item \textbf{Knowledge Probes:} Knowledge probes are structured queries used to examine how well a model represents specific factual or conceptual knowledge. These probes help determine whether the model has not only memorized but also semantically understood the knowledge during training. Two types of test sets are constructed for comprehensive evaluation: a fact-based test set and a concept-based test set.
    \item \textbf{Query Optimization with Leading Text:} To improve the model's ability to retrieve relevant knowledge from its internal representations, we introduce "leading text"—a form of prompt engineering where additional contextual cues are prepended to the query. For example: "Based on the medical knowledge you have learned, what is the most likely diagnosis for a patient presenting with..."
\end{itemize}

The introduction of knowledge injection significantly improves the model’s performance on both fact-based and concept-based knowledge probes. Specifically, accuracy increases from 39\% to 60.57\%, indicating a substantial enhancement in the model’s ability to access and reason over the injected knowledge. The improvement is particularly notable in concept-based tasks, suggesting that the model has developed a better-structured understanding of medical domains. These results demonstrate that knowledge injection, when effectively integrated and evaluated through targeted probing methods, can significantly enhance the reasoning and knowledge utilization capabilities of large medical language models.

\subsection{Medical Records}
Real-world, high-quality medical records are invaluable for training medical foundation models, providing authentic clinical narratives, diagnostic reasoning, and treatment plans rarely captured in textbooks or guidelines. In this work, we curate a large-scale corpus from two practice-proximal channels: public online medical consultation dialogues and de-identified electronic health records (EHRs) released via public repositories. For confidentiality and compliance, we do not disclose exact dataset names or volumes; only aggregated, de-identified, privacy-filtered text is retained for modeling.

\paragraph{Online Medical Consultations}
We collect publicly available consultation dialogues that reflect symptom narratives, clinician questioning, preliminary differentials, and triage or follow-up suggestions. These short- and medium-form interactions complement formal records by capturing colloquial descriptions, lay terminology, and pragmatic decision cues encountered in routine care.

\paragraph{Public EHR Collections}
We draw on de-identified EHR datasets made publicly accessible through established releases, spanning both outpatient and inpatient encounters. Exact counts are intentionally abstracted (reported internally only) to reduce re-identification risk; qualitatively, coverage spans common ambulatory presentations through complex inpatient trajectories.
\begin{itemize}
\item \textbf{Outpatient EHRs:} Emphasize common presentations (chief complaint, history of present illness, assessment/plan, prescriptions) with broad breadth rather than disclosing absolute volume.
\item \textbf{Inpatient EHRs:} Include admission notes, longitudinal progress notes, procedure and operative narratives, discharge summaries, laboratory panels, and imaging impressions across a wide diagnostic mix.
\end{itemize}
All materials—whether originally released in de-identified form or derived from public dialogues—undergo a conservative PHI-removal pipeline. We further normalize and segment unstructured text into coherent clinical documents. Quality is enforced through automatic discriminator models and physician-led spot audits. The corpus is used for continued pre-training to capture the structure and lexicon of clinical documentation, and for supervised fine-tuning (SFT) to strengthen the model’s reasoning in complex scenarios.

\section{Method}
\label{sec:method}

This section details the multi-stage training methodology for QuarkMed, which includes Instruction Fine-Tuning (IFT), Supervised Fine-Tuning (SFT), and two distinct stages of Reinforcement Learning (RL) designed to instill specialized reasoning and general alignment.

\subsection{Instruction Fine-Tuning (IFT)}
\label{subsec:ift}

The initial phase of our training pipeline is Instruction Fine-Tuning (IFT), a critical step to align the general-purpose, pre-trained base model with the ability to follow instructions within the specialized medical domain. While pre-trained language models (PLMs) acquire extensive world knowledge through their auto-regressive training objective, they often fail to adhere to user-specific directives, a phenomenon known as the "alignment gap" \cite{ouyang2022traininglanguagemodelsfollow, azizi2022medpalm}. IFT seeks to bridge this gap by fine-tuning the model on a large and diverse dataset of instruction-response pairs. This process transforms the model from a simple text completion engine into a capable assistant that can understand and execute professional medical tasks. To this end, we have systematically constructed a task-oriented dataset comprising hundreds core IFT tasks with over 400,000 high-quality samples, following the principles of multitask-prompted training \cite{sanh2022multitaskpromptedtrainingenables, wang2022supernaturalinstructionsgeneralizationdeclarativeinstructions}.

\begin{figure}
    \centering
    \includegraphics[width=0.8\linewidth]{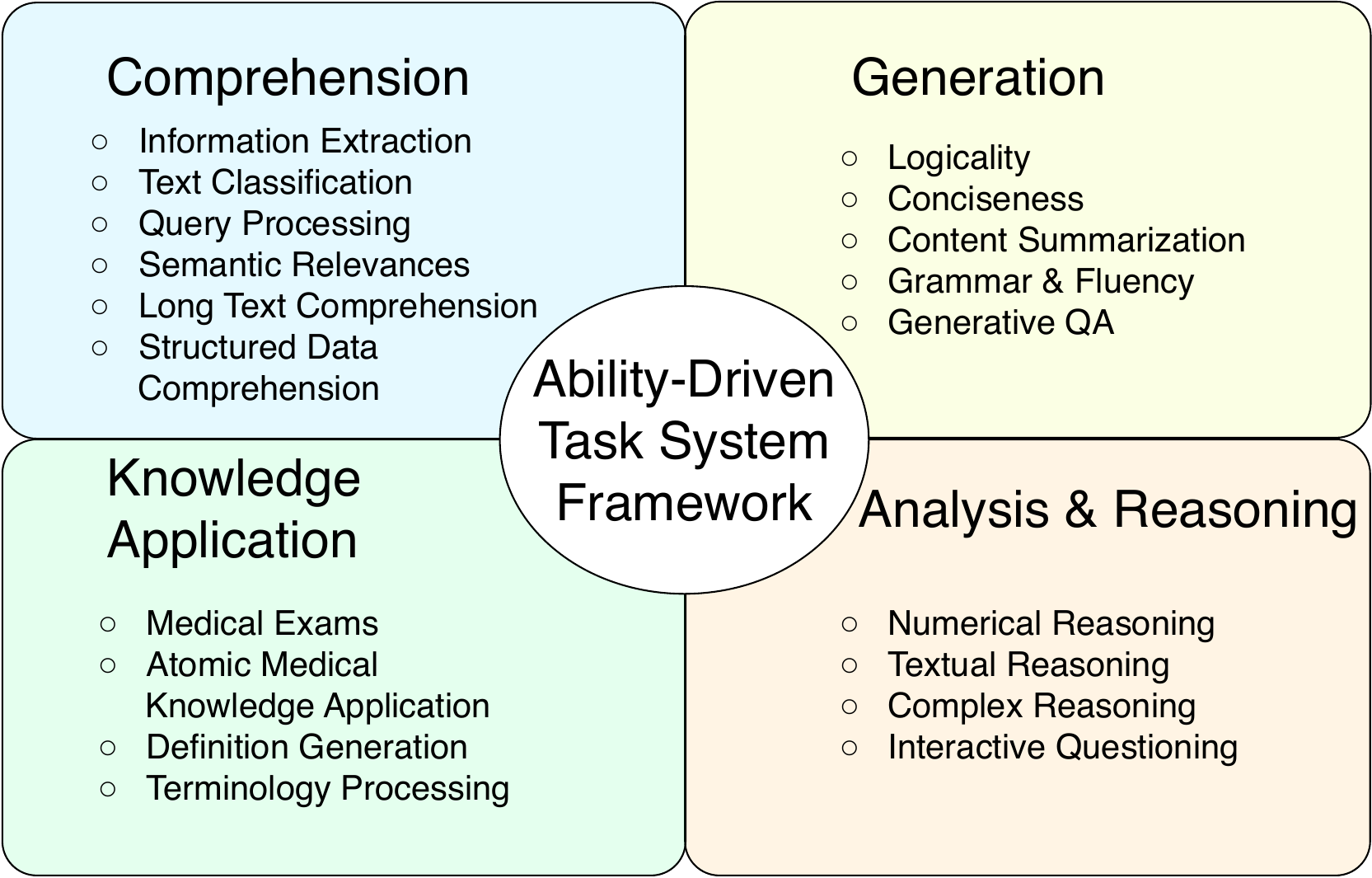}
    \caption{Ability driven data augmentation loop}
    \label{fig:ability_driven}
\end{figure}

\paragraph{Task Design Framework}
To construct a comprehensive instruction set, we adopted a dual-pronged strategy: a foundational, \textbf{Ability-Driven} framework and a responsive, \textbf{Problem-Driven} augmentation loop. This approach ensures broad capability coverage while systematically addressing identified model weaknesses.
The Ability-Driven framework, illustrated in \cref{fig:ability_driven}, deconstructs the requirements of a medical AI assistant into a four-dimensional capability system.
\begin{itemize}
\item \textbf{Comprehension:} This dimension targets the model's fundamental understanding of prompts, including instructions, queries, and contextual information. We treat traditional Natural Language Understanding (NLU) tasks (e.g., information extraction, text classification, semantic similarity) as atomic abilities, which form the bedrock for performing more complex, domain-specific tasks.
\item \textbf{Generation:} To ensure logical coherence, conciseness, and fluency, we designed tasks to refine specific aspects of text generation. For instance, a "Sentence Ordering" task enhances logical flow, while tasks for discriminating between hyponyms/hypernyms and synonyms reduce redundancy in enumerated lists.
\item \textbf{Knowledge Application:} This dimension aims to activate the model's capacity to apply its parametric medical knowledge appropriately in different contexts. Tasks include adjusting terminology for diverse audiences (e.g., clinical vs. layperson) or applying knowledge of contraindications for specific populations (e.g., pediatric or geriatric patients) \cite{singhal2022largelanguagemodelsencode}.
\item \textbf{Analysis and Reasoning:} As the cornerstone of advanced medical AI, this dimension focuses on multi-step reasoning. We constructed a curriculum of reasoning tasks, beginning with atomic reasoning skills such as unit conversion and numerical comparison \cite{cobbe2021trainingverifierssolvemath}, and progressing to complex reasoning scenarios like multi-turn diagnostic dialogues and inference from clinical notes, inspired by chain-of-thought methodologies \cite{wei2023chainofthoughtpromptingelicitsreasoning}.
\end{itemize}

\begin{figure}
    \centering
    \includegraphics[width=0.4\linewidth]{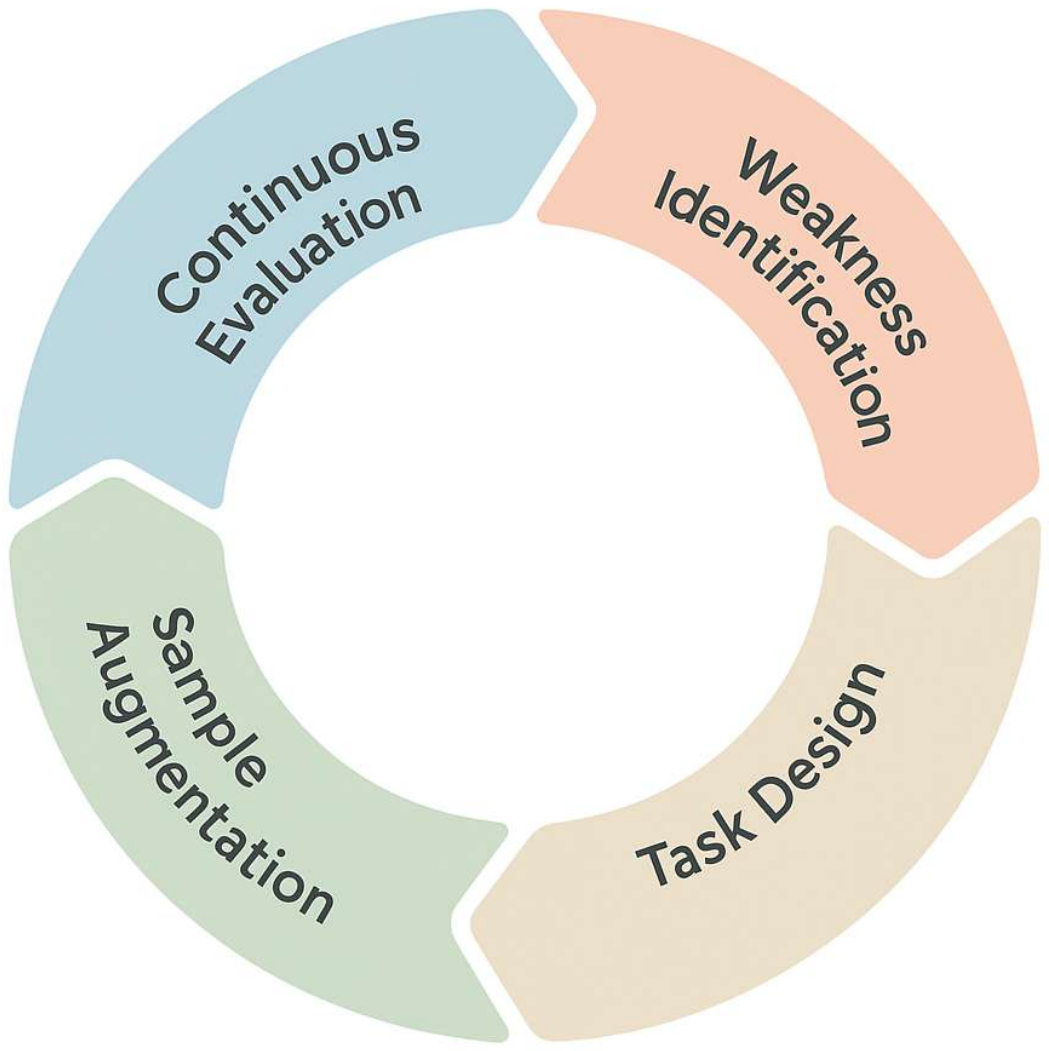}
    \caption{Problem driven data augmentation loop}
    \label{fig:problem_driven}
\end{figure}

The Problem-Driven strategy establishes a continuous optimization cycle to address specific performance deficits identified during evaluation. As depicted in \cref{fig:problem_driven}, this iterative loop involves identifying model weaknesses, designing targeted tasks, and augmenting the training data to address these gaps. Key examples include:
\begin{itemize}
\item \textbf{Counterfactual Robustness:} To mitigate factual hallucinations, a "Factual Consistency Judgment" task was designed to train the model to identify and refuse to answer questions based on false premises.
\item \textbf{Output Stability:} To improve robustness to linguistic variations, we generated "synonymous instruction - same output" pairs, ensuring the model produces consistent responses to semantically equivalent queries.
\item \textbf{RAG Noise Resistance:} To enhance performance in Retrieval-Augmented Generation (RAG) scenarios \cite{lewis2021retrievalaugmentedgenerationknowledgeintensivenlp}, we constructed noisy samples containing both relevant and irrelevant retrieved passages. This trains the model to accurately identify, cite, and synthesize information from the most pertinent sources while ignoring distractors.
\end{itemize}
\paragraph{Guiding Principles for Task Construction}
The construction of our IFT dataset was governed by three core principles to ensure its effectiveness.
\begin{itemize}
\item \textbf{Task Atomicity:} Each task was designed to target a single, well-defined objective. This principle facilitates precise capability tracking and simplifies the attribution of performance changes, making the optimization process more controllable.
\item \textbf{Instruction Generalization:} We developed unique prompt templates for each IFT task. This approach encourages the model to learn the underlying instruction-following behavior rather than memorizing surface-level patterns, thereby enhancing generalization to unseen tasks and isolating these foundational abilities from downstream, application-specific SFT.
\item \textbf{Task Decomposition:} Complex, multi-step tasks that are difficult to learn end-to-end were systematically decomposed into more tractable sub-tasks. For example, in RAG scenarios, we created a distinct "Relevance Extraction" task to train the model to first identify useful information before generating a final answer.
\end{itemize}
\paragraph{Data Sourcing and Sample Construction}
We employed a multi-faceted strategy to generate high-quality training data tailored to the demands of each task.
\begin{itemize}
\item \textbf{High-Quality Base Samples:} For foundational NLU and generation tasks, we established a gold-standard reference by sampling outputs from multiple large models, followed by a cross-validation and voting process, and concluding with manual verification by domain experts.
\item \textbf{Complex and Adversarial Samples:} For challenging tasks such as complex reasoning and counterfactual handling, we utilized a Self-Instruct approach \cite{wang2023selfinstructaligninglanguagemodels}, providing few-shot exemplars to guide a large model in generating a diverse set of novel prompts and responses.
\item \textbf{Safety Alignment Samples:} To improve robustness against misuse, we trained a dedicated "Red-Teaming" model to generate adversarial prompts, enabling us to fine-tune the model for safer and more harmless responses, in line with established safety protocols \cite{ganguli2022redteaminglanguagemodels, bai2022traininghelpfulharmlessassistant}.
\end{itemize}
\paragraph{Training Strategy}
The IFT phase employed a sophisticated training strategy to maximize efficiency and model performance.
\begin{itemize}
\item \textbf{Curriculum Learning:} Recognizing that tasks exhibit varying levels of difficulty and that certain abilities are prerequisites for others (e.g., information extraction precedes summarization), we adopted a curriculum learning strategy \cite{10.1145/1553374.1553380}. Training progressed from simpler, atomic tasks to more complex, composite ones, which improved both convergence speed and final model performance.
\item \textbf{Task Sample Ratio Optimization:} A key challenge is determining the optimal sampling ratio across the 112 tasks to achieve a balanced set of capabilities. We addressed this using a data-driven optimization process. First, we established automated evaluation suites for each core ability. We then used Bayesian Optimization, modeling the relationship between inter-group sampling ratios and a weighted overall performance score with a Gaussian Process Regression (GPR) model. An infill criterion was used to efficiently explore the high-dimensional search space and identify a near-optimal ratio distribution \cite{FORRESTER200950}.
\end{itemize}
Through this systematic IFT process, the model's foundational abilities and instruction-following fidelity were significantly enhanced, providing a robust starting point for subsequent stages of fine-tuning and reinforcement learning.

\subsection{Supervised Fine-Tuning (SFT)}
\label{subsec:rft}

\begin{figure}[htp]
    \centering
    \includegraphics[width=0.8\linewidth]{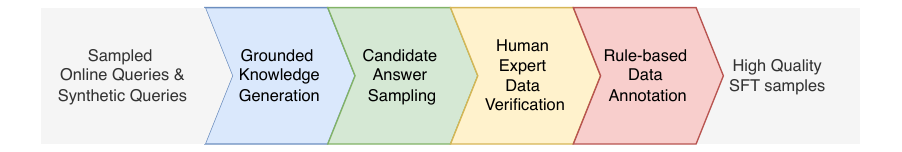}
    \caption{Data processing pipeline for SFT samples}
    \label{fig:data_curation}
\end{figure}

The development of a safe, accurate, and helpful medical large language model hinges on the quality of its Supervised Fine-Tuning (SFT) data. To this end, we have designed a meticulous data processing pipeline that ensures our training samples are diverse, robust, and medically sound. The general process for the high-quality SFT samples generation is shown in \cref{fig:data_curation}.

\paragraph{Query Selection Process} Our strategy for Supervised Fine-Tuning (SFT) is built on a hybrid approach, utilizing both synthetic and real-world online data to create a medically proficient and adaptable AI. The primary goal is to develop a comprehensive set of skills, ranging from fundamental comprehension to advanced clinical reasoning. Synthetic data is systematically generated to train and test the model across a wide spectrum of controlled scenarios, establishing a strong foundation of core capabilities. This is complemented by data from online sources, which attunes the model to the nuance and diversity of real-world user queries, ensuring its practical applicability.

A significant focus of our training is on cultivating advanced comprehension and reasoning abilities. The model is trained to summarize lengthy and complex documents, synthesize key information scattered across multiple texts, and perform logical inference to connect disparate points. This training also involves handling information from diverse genres, such as clinical guidelines, Q\&A forums, and patient dialogues. Furthermore, the model is taught to resolve conflicts by selecting the most accurate information from contradictory sources and to integrate knowledge from external tools, like BMI calculators, to provide more comprehensive answers. \Cref{tab:sft_key_capabilities} shows the key capabilities for the synthetic data categories.

Ensuring safety and accuracy is paramount, and the model undergoes rigorous training to enhance its robustness against erroneous or misleading information. It is specifically trained to identify and resist various forms of disturbance, such as ignoring irrelevant reference materials, differentiating between semantically similar but incorrect medical concepts, and flagging factually incorrect information. This error resistance extends to the user's query itself, enabling the model to recognize and handle questions that contain flawed premises, factual inconsistencies, or irrational assumptions, thereby preventing the generation of unsafe or nonsensical responses.

\begin{table}[h!]
\centering
\caption{Key Capabilities for SFT Synthetic Data Strategy}
\label{tab:sft_key_capabilities}
\renewcommand{\arraystretch}{1.2} 
\begin{tabular}{lp{5cm}p{5cm}}
\toprule
\textbf{Capability Category} & \textbf{Specific Capability} & \textbf{Purpose} \\
\midrule
\textbf{Summarization \& Induction} 
& Dispersed Information Synthesis & To combine scattered information into a coherent answer. \\
\midrule
\textbf{Disturbance Resistance} 
& Contradictory Information & To identify and use correct information from conflicting sources. \\
\midrule
\textbf{Error Resistance} 
& Factual Inconsistency & To recognize and handle incorrect or illogical user queries. \\
\midrule
\textbf{Fundamental Capabilities} 
& Timeliness & To provide the most current and updated information. \\
& Authoritativeness & To learn to prioritize and cite authoritative sources. \\
\bottomrule
\end{tabular}
\end{table}

Finally, this training is grounded by a focus on foundational skills and real-world performance. The model is explicitly taught to ensure its responses are logically structured, use the most current and timely information, and cite authoritative sources to build user trust. Capabilities such as correcting misspelled drug names and maintaining coherent, non-repetitive conversational flow are also instilled. By incorporating online data from clinical and general health contexts, we ensure these foundational skills are effectively applied to address genuine user needs, from complex clinical questions to practical advice on healthy living, while upholding strict safety standards against adversarial inputs.

\paragraph{Data Curation Process} Our data curation process consists of four main stages: medical knowledge grounded generation, candidate answer sampling, human-expert data verification, and rule-based data annotation. 

\begin{itemize}
    \item \textbf{Medical Knowledge Grounded Generation}: To generate responses grounded in medical knowledge, we begin by sampling anonymous, real-world online queries. We leverage the powerful retrieval capabilities of Quark Medical Search to gather a comprehensive set of reference materials. This includes professional medical literature and medical question-and-answer forums, supplemented with timely, web-wide content such as informational notes. The objective is to create a SFT dataset that reflects real-world complexity, possesses a sufficient level of difficulty, and covers a wide spectrum of scenarios across medical knowledge, clinical practice, and medical application. All original data sources, such as proprietary medical texts and patient records, undergo rigorous privacy protection measures.
    \item \textbf{Candidate Answer Sampling}: For each prompt, we generate multiple candidate answers tailored to the specific requirements of each data sample. Utilizing a proprietary medical quality model alongside the capabilities of in-house reward models, we employ a "Best-of-N" strategy to select the most optimal response to serve as a candidate for the next stage.
    \item \textbf{Human-Expert Data Verification}: We assemble a team of medical experts to meticulously review and refine the candidate answers. This verification process ensures that every response adheres to our stringent standards of safety, accuracy, and usefulness. To enhance the quality and efficiency of this stage, annotators follow a structured format that includes the reference materials, the initial question, the model's best-selected answer, responses from other major models, and a summary of key points potentially required in the final answer.
    \item \textbf{Rule-Based Data Annotation}: In the final stage, we apply rule-based data annotations. This involves real-time services that provide automated validation of formatting and correctness, further elevating the overall quality and consistency of the annotated data.
\end{itemize}

Ultimately, this comprehensive process, which combines systematic query design with a multi-stage curation pipeline leveraging both automated systems and human medical expertise, ensures the creation of high-quality SFT data. This meticulous approach is fundamental to developing a trustworthy and capable medical AI assistant that is both safe and genuinely helpful to users.

\subsection{Stage 1 RL: Large-Scale Medical Reinforcement Learning}
\label{subsec:r0_training}
Fields like medicine—especially core tasks such as disease diagnosis, rational drug use, and test ordering—are inherently knowledge-intensive and rely heavily on sophisticated reasoning. To significantly elevate the QuarkMed model's reasoning capabilities in these complex scenarios, we implemented a specialized Reinforcement Learning (RL) phase focused exclusively on reasoning-based tasks. By using a multi-task learning approach across medical board exams, disease diagnosis, appropriate medication prescribing, and lab/imaging test selection, our goal is to systematically enhance the model's overall medical reasoning competence.

\paragraph{Model Initialization with SFT}
To accelerate RL convergence and conserve computational resources, we begin with a "cold-start" strategy using Supervised Fine-Tuning (SFT). The objective is to ensure the model acquires a baseline reasoning capability and can adhere to predefined formats (e.g., JSON) before entering the more complex RL stage. We fine-tune the base model directly on a curated set of over 700 high-quality annotated examples from our target reasoning tasks. To preserve model diversity and prevent its entropy from dropping too low, we limit SFT to just two epochs. This encourages the model to "learn to reason" rather than "memorize answers," building a strong foundation for the exploration required in RL.

\paragraph{High-Quality Data Pipeline for RL Training}
The success of reinforcement learning hinges on the quality of its training data. Our data pipeline was designed around three core principles: diversity, difficulty, and accuracy. For \textbf{diversity}, we start with heterogeneous data sources, including electronic health records and medical exam questions, and use a label-based stratified sampling strategy to ensure the training data is balanced. For \textbf{difficulty}, we built a dynamic, model-aware filter to continuously feed the model challenging examples by screening for high-accuracy samples, analyzing reasoning complexity, and synthesizing more complex problems. This filtering pipeline was customized for each model architecture (Llama vs. Qwen) and size (8B vs. 32B). For \textbf{accuracy}, we use a "model-assisted, expert-verified" workflow where a high-performing model generates initial responses, discrepancies are flagged, and our team of medical experts reviews and corrects them to guarantee label accuracy.

\begin{figure}[h]
    \centering
    \includegraphics[width=0.5\linewidth]{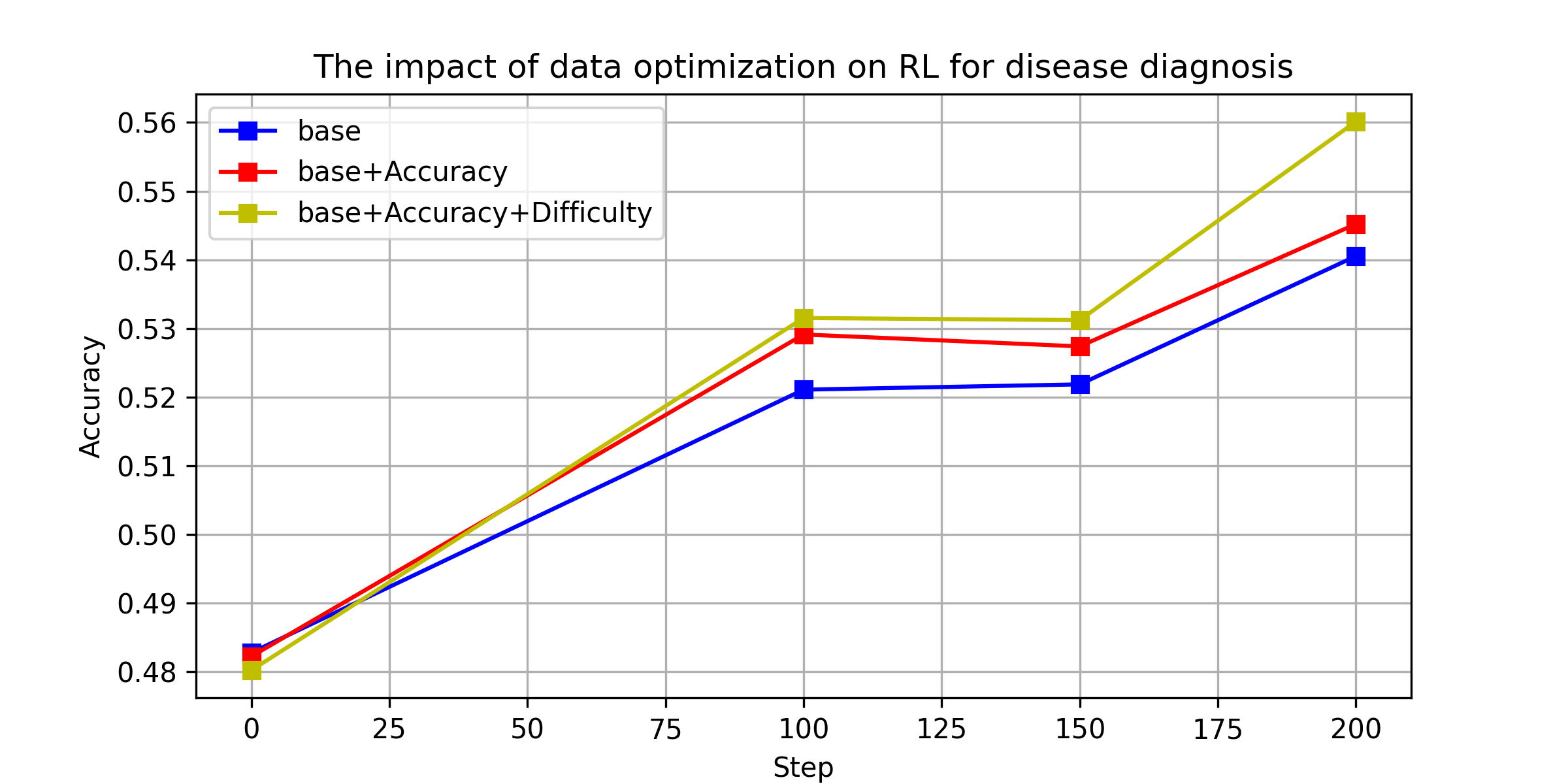}
    \caption{This figure illustrates the impact of enhancing data accuracy and difficulty on RL performance. Both factors were found to positively influence the model's effectiveness in disease diagnosis tasks.}
    \label{fig:rl_difficulty}
\end{figure}

\paragraph{Reward Model Design}
A precise reward signal is critical for guiding the RL training process. For reasoning tasks with a clear ground truth, we designed a hybrid reward model, encapsulated in a "Verifier," that prioritizes rules but is augmented by a model-based component. This approach is founded on a rule-first principle, where objective and stable reward signals from established medical rules are given precedence to prevent reward hacking. We also standardized output formats (e.g., ICD codes for diagnosis) to enable automated scoring. However, since simple rule-based matching can be brittle, we introduced a model-based reward to handle synonyms, hierarchies, and incomplete labels. To mitigate potential bias, this Verifier was iteratively optimized. Experiments confirmed that this hybrid "rule + model" strategy improved disease diagnosis performance by 3 percentage points over a purely rule-based method. Finally, a dedicated format-adherence reward is included to ensure outputs strictly follow the required structure. The training data and Verifier setup are summarized in the table below.

\begin{table}[h]
\centering
\caption{Task-Specific Verifier Composition and Key Metrics}
\renewcommand{\arraystretch}{1.2}
\begin{tabular}{@{}llll@{}}
\toprule
\textbf{Task Type} & \textbf{Data Volume} & \textbf{Verifier Composition} & \textbf{Key Metrics} \\ \midrule
Disease Diagnosis & 16,000 & Rules (ICD Matching) + Model & Accuracy, Recall \\
Rational Drug Use & 10,000 & Rules (JSON Extraction) + Model & Drug Entity Accuracy \\
Test Ordering & 16,000 & Rules (Keyword Matching) & Accuracy of Recommendations \\
Medical Exam Questions & 27,000 & Rules (Exact Answer Match) & Answer Accuracy \\ \bottomrule
\end{tabular}
\end{table}

\paragraph{RL Implementation and Optimization}
We used the VeRL \cite{Sheng_2025} framework for RL training, selecting the Group Relative Policy Optimization (GRPO) algorithm. GRPO normalizes the advantage function within groups of samples, which naturally supports multi-task training and improves stability. For efficiency and stability, we implemented several optimizations. First, dynamic resampling at the start of each epoch removes simple samples the model has already mastered, boosting training efficiency by about 20\%. Second, to improve the stability of the policy model's exploration and the accuracy of the Verifier's scores, we increased the number of rollouts to 32 for the policy model and 8 for the Verifier. The final stage 1 model's performance on our test sets is shown below, demonstrating substantial improvements over the baseline.

\begin{table}[h]
\centering
\caption{Stage 1 Model Performance Comparison}
\renewcommand{\arraystretch}{1.2}
\begin{tabular}{@{}lcccccc@{}}
\toprule
\multirow{2}{*}{\textbf{Model}} & \multicolumn{4}{c}{\textbf{Chinese National Medical Licensing Examination}} & \multicolumn{2}{c}{\textbf{Disease Diagnosis}} \\ \cmidrule(lr){2-5} \cmidrule(lr){6-7}
 & \textbf{Junior} & \textbf{Intermediate} & \textbf{Assoc. Senior} & \textbf{Senior} & \textbf{Top-1 Acc.} & \textbf{List Score} \\ \midrule
DeepSeek-R1 & 0.814 & 0.826 & 0.723 & 0.387 & 0.75 & 1.46 \\
Quark Stage1 & 0.822 & 0.772 & 0.683 & 0.524 & 0.86 & 3.32 \\ \bottomrule
\end{tabular}
\end{table}

\subsection{Stage 2 RL: General Reinforcement Learning Integration}
\label{subsec:r1_rl_training}
The primary objective of the general RL stage is to employ Reinforcement Learning (RL) to align the model's behavior with human preferences and values. This process involves developing a Reward Model (RM) to quantitatively assess the quality of model outputs and implementing an RL algorithm to optimize the policy based on these reward signals. To ensure the model generates high-quality medical responses, our RM holistically evaluates model outputs across three core dimensions: Honesty, Helpfulness, and Content Compliance.

\begin{figure}[h]
    \centering
    \includegraphics[width=0.75\linewidth]{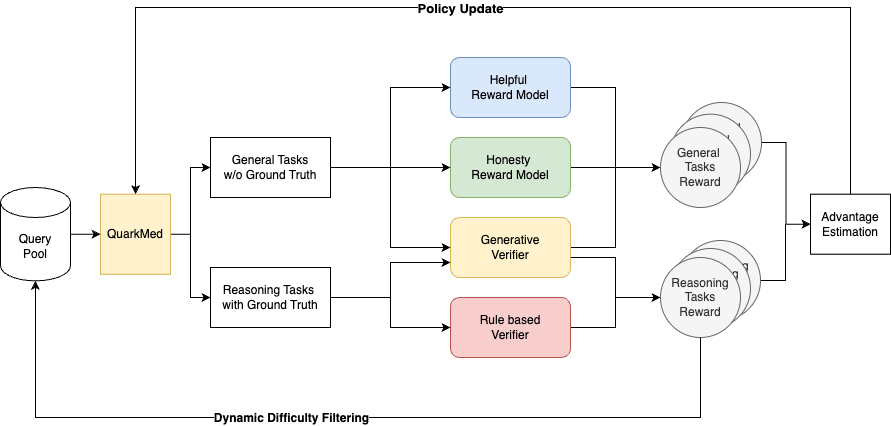}
    \caption{Overall training method for the general RL stage. Three types of reward signals are used for general and reasoning tasks during the training.}
    \label{fig:enter-label}
\end{figure}



\begin{figure}[h]
    \centering
    \includegraphics[width=0.9\linewidth]{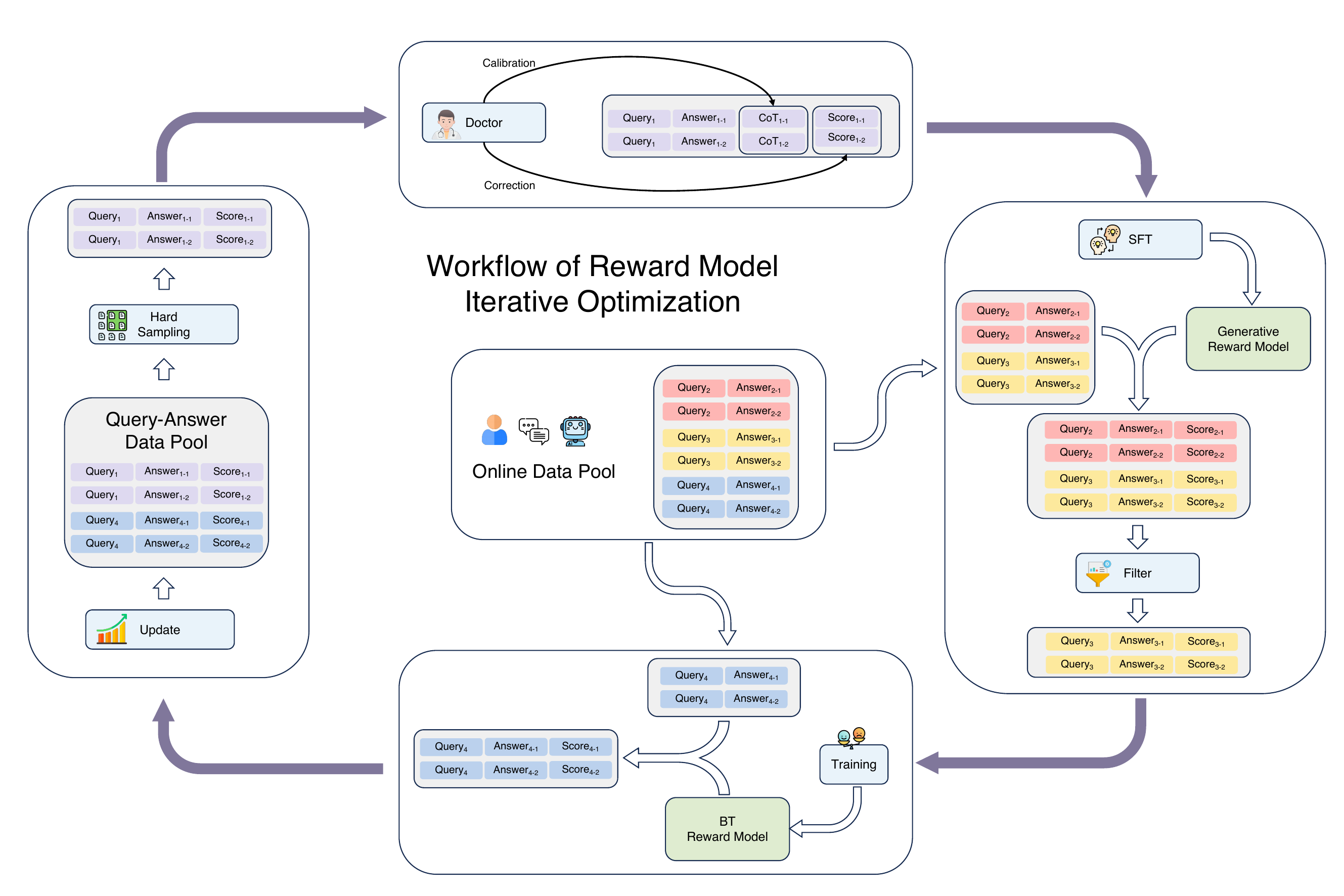}
    \caption{Workflow of reward model iterative optimization. An iterative optimization loop that begins with manually labeled data to train a generative chain‑of‑thought (CoT) reward model. The CoT‑reward model is used for pairwise selection to obtain high‑quality pairs, which are then used to train a BT‑reward model. The BT‑reward model is further employed in an active learning setting to continuously improve both the training samples and the performance of the reward models.}
    \label{fig:iterative_reward}
\end{figure}

\paragraph{Honesty Reward}
The Honesty Reward is designed to ensure the medical accuracy of the model's responses. To address the challenge of costly manual annotation for factual errors, we designed an iterative optimization loop involving a generative reward model model and the reward model, as show in \cref{fig:iterative_reward}. First, we trained a generative reward model on manually calibrated SFT samples with Chain-of-Thought (CoT) reasoning and score. This model was then used to score multiple candidate responses and generate high-quality preference pairs. Combining these filtered pairs with manually annotated samples, we trained the Reward Model (RM) using the Bradley-Terry (BT) model \cite{bradley1952rank}. Finally, in a closed-loop iteration, the new RM evaluated more candidates, and samples with erroneous or ambiguous scores were fed back to human annotators for re-labeling, continuously enhancing both models.

\paragraph{Helpfulness Reward}
The evaluation criteria for helpfulness align closely with human intuition, making annotation relatively straightforward. We sampled tens of thousands of user prompts from online logs and used multiple models to generate diverse candidate responses. Annotators then provided preference labels on these varied outputs, which enhanced the generalization ability of the reward model. To counter "reward hacking" \cite{gao2022scalinglawsrewardmodel}, where the policy learns to game the RM (e.g., by inflating response length), we established a continuous feedback loop, re-labeling new samples generated during RL training to iteratively update the RM and enhance its robustness.

\paragraph{Consistency Reward}
To address the issue of inconsistency between the reasoning process and the final summary in the model's outputs, we developed a dedicated consistency reward model. In order to improve model accuracy, we have constructed a multi-stage data iteration pipeline to enhance data quality. This pipeline incorporates key stages such as automated data collection via large model auto-labeling, model-based label calibration, and iterative active learning. Samples refined through this multi-stage process achieve a consistency reward score of over 80.

\paragraph{General Verifier for Content Compliance}
To meet strict style and formatting requirements for specific scenarios (e.g., health notes), we trained a General Verifier, which is an instruction-following model. By providing it with explicit evaluation principles, it can score responses based on adherence to these rules. This approach is highly flexible, as it can be adapted to new standards by modifying its guiding principles. It also effectively mitigates hacking, as new validation rules can be quickly added to address stylistic issues that are difficult for traditional RMs to cover.

\paragraph{Training Data Construction}
The training data for the RL stage comprises approximately 80,000 prompts, divided into reasoning-intensive and general-purpose categories. The reasoning-intensive data (approx. 60,000 prompts) was isolated from data used in earlier stages to ensure continued gains. The general-purpose data (approx. 20,000 prompts), sourced from online logs, was selected via diversity sampling. We used the SFT-stage model to perform multiple rollouts and prioritized prompts that yielded a high diversity in reward scores, indicating varied quality in model responses. Ultimately, we maintained a reasoning-to-general-purpose data ratio of approximately 3:1.

\paragraph{Algorithm Comparison and Selection}
We conducted a comparative analysis of DPO (Direct Preference Optimization) \cite{rafailov2024directpreferenceoptimizationlanguage} and GRPO (Group Relative Policy Optimization) to evaluate their effectiveness. For DPO, we formed preference pairs from the highest- and lowest-rewarded responses for each prompt. For GRPO, we sampled 8 candidate responses per prompt and applied the appropriate reward models or the General Verifier for scoring, with a KL divergence penalty coefficient of 0.01. Experimental results, shown in \cref{tab:rl_algo_comparison}, demonstrated that GRPO significantly outperformed DPO across most dimensions. The GRPO model achieved the best performance in the overall score and on key dimensions such as Honesty and Harmlessness. Consequently, we selected GRPO as the core RL algorithm for this project.

\newcolumntype{C}{>{\centering\arraybackslash}X}

\begin{table}[h]
\centering
\caption{Performance Comparison of RL Algorithms}
\label{tab:rl_algo_comparison}
\renewcommand{\arraystretch}{1.2} 
\begin{tabularx}{\textwidth}{@{} l *{7}{C} @{}}
\toprule
\textbf{Model / Algorithm} & \textbf{Overall \mbox{(5-pt)}} & \textbf{Honesty \mbox{(3-pt)}} & \textbf{Harmless \mbox{(3-pt)}} & \textbf{Logicality \mbox{(3-pt)}} & \textbf{Formatting \mbox{(3-pt)}} & \textbf{Relevance \mbox{(3-pt)}} & \textbf{Comprehensive \mbox{(3-pt)}} \\
\midrule
DeepSeek-R1 (Baseline) & 3.60 & 2.24 & 2.76 & 2.80 & 2.92 & 2.94 & 2.60 \\
DPO & 3.49 & 2.16 & 2.72 & 2.66 & 2.88 & 2.92 & 2.72 \\
GRPO & 3.84 & 2.40 & 2.88 & 2.82 & 2.94 & 2.94 & 2.56 \\
\bottomrule
\end{tabularx}
\end{table}

\section{Evaluation}
\label{sec:evaluation}

We conducted a comprehensive evaluation of the QuarkMed model using a wide range of benchmarks, including prominent open-source medical evaluation suites and several private datasets. We categorized the medical tasks into three main types: Medical Question Answering (QA), Medical Reasoning, and Foundational Medical Capabilities. \cref{tab:open_bench_table,tab:internal_bench_table} presents a summary of the results across these tasks and datasets, with detailed descriptions of each provided in the subsequent sections.

\subsection{Evaluation Methodology}
\label{subsec:evaluation_methodology}

Due to data privacy and licensing constraints, all competitor models selected for comparison were either open-source models or accessible via publicly available APIs. During the evaluation, we performed a single inference pass for each test sample. Across all benchmark tests, a temperature parameter of 0.6 was consistently used for inference with every model. 
For certain datasets, such as MMLU \cite{hendryckstest2021}, where test set answers are not provided, we conducted our evaluation on the validation set. For datasets with an excessively large number of test samples, such as RareBench \cite{10.1145/3637528.3671576}, we performed uniform sampling, using a maximum of 1,000 samples per test set. We re-executed the entire evaluation pipeline for all selected models and APIs using a standardized prompt to ensure consistent results.
For multiple-choice questions, models were prompted to provide the final answer in JSON format. For open-ended question formats, we employed DeepSeek-V3-0324 \cite{deepseekai2024deepseekv3technicalreport} for standardized post-processing and answer scoring.

\subsection{Datasets}
\label{subsec:datasets}

\paragraph{Medical Question Answering} This category includes datasets such as MedQA \cite{jin2020diseasedoespatienthave}, MedMCQA \cite{pmlr-v174-pal22a}, PubMedQA \cite{jin2019pubmedqa}, CMExam \cite{liu2023benchmarking}, and AfriMed-QA \cite{olatunji2024afrimed}.
\begin{itemize}
    \item \textbf{MedQA}: is an open-ended multiple-choice question dataset for the medical domain. In this paper, we only used the USMLE section data.
    \item \textbf{MedMCQA}: contains high-quality multiple-choice questions from the AIIMS and NEET PG entrance examinations, covering 2,400 medical topics and 21 medical subjects.
    \item \textbf{PubMedQA}: is a biomedical question-answering dataset constructed from PubMed abstracts. Given a medical research question, the model must answer "yes," "no," or "maybe" based on the provided abstract.
    \item \textbf{CMExam}: is derived from the Chinese National Medical Licensing Examination and contains over 60,000 multiple-choice questions for standardized medical assessment.
    \item \textbf{AfriMed-QA}: is a large-scale, open-source dataset featuring clinically diverse questions and answers from a Pan-African context. It is designed for the rigorous evaluation of large language models on accuracy, factualness, hallucination, demographic bias, potential harms, comprehension, and memory. We utilized only the multiple-choice portion of this dataset.
\end{itemize}

\paragraph{Medical Reasoning} This category comprises datasets focused on disease reasoning and diagnosis, including DiagnosisArena \cite{zhu2025diagnosisarena}, RareBench, MedXpertQA\cite{zuo2025medxpertqa}, and others.
\begin{itemize}
    \item \textbf{MedXpertQA} is a highly challenging and comprehensive benchmark designed to evaluate expert-level medical knowledge and advanced reasoning capabilities. 
    \item \textbf{DiagnosisArena} contains 1,000 paired, multi-turn patient case dialogues and their corresponding diagnoses, designed to evaluate the diagnostic reasoning capabilities of large language models in a clinical setting.
    \item \textbf{RareBench} is used to systematically evaluate the capabilities of large language models across four key dimensions within the rare disease domain. We used only the differential diagnosis section.
    \item \textbf{MedBullets} \cite{chen-etal-2025-benchmarking} is a medical question bank of simulated clinical problems, including 308 clinical multiple-choice questions in the style of USMLE Step 2 and Step 3.
    \item \textbf{CMB-clin} \cite{wang2024cmbcomprehensivemedicalbenchmark} based on complex real-world clinical cases, assesses a model's ability to apply knowledge in authentic diagnostic and treatment scenarios, evaluating whether it can leverage this knowledge to solve complex clinical problems.
    \item \textbf{CPQExam} is a private dataset built from the Chinese Health Professional Qualification Examination, an exam that physicians in China are required to pass for professional licensure and career advancement. The exam contains a large number of questions focusing on case analysis and practical application.
    \item \textbf{MediQ} \cite{li2024mediq} simulates interactive dialogues between patients and specialists. We used only the diagnosis-related multiple-choice questions, which require deriving a final diagnosis from the provided information.
    \item \textbf{RedisQA}\cite{wang2024assessing} is a dataset built around rare disease diagnosis, covering 205 rare diseases to evaluate the performance of large language models in this area.
\end{itemize}

\paragraph{Foundational Medical Capabilities} This category consists of the medical-related subsets of MMLU and the MedCalc \cite{NEURIPS2024_99e81750} benchmark.
\begin{itemize}
    \item \textbf{MMLU} is a multi-task benchmark featuring multiple-choice questions from various fields of knowledge. We selected only its medical-related subjects: virology, professional medicine, medical genetics, college medicine, college biology, clinical knowledge, and anatomy.
    \item \textbf{MedCalc} is a benchmark designed to assess a model's proficiency as a clinical calculator. Each sample includes a question that requires the model to calculate a clinical value based on a provided patient note.
\end{itemize}
\begin{table*}[htbp]
  \centering
  \caption{Performance on Open Benchmarks.}
  \label{tab:open_bench_table}
  \resizebox{\textwidth}{!}{
  \begin{tabular}{lp{1.5cm}p{1.5cm}p{1.5cm}p{1.5cm}p{1.5cm}p{1.5cm}|p{1.5cm}p{1.5cm}p{1.5cm}p{1.5cm}}
    \toprule
    & \multicolumn{6}{c}{\textbf{Large Models}} & \multicolumn{4}{c}{\textbf{Medium-sized Models}} \\
    \cmidrule(lr){2-7} \cmidrule(lr){8-11}
    \textbf{Dataset} & 
    \textbf{Gemini-2.5-pro-0617} &
    \textbf{gpt-4o-2024-08-06} &
    \textbf{DeepSeek-R1-0528} & %
    \textbf{Kimi-k2} & 
    \textbf{Qwen3-235B-A22B-Instruct-2507} &
    \textbf{Qwen3-235B-A22B} & %
    \textbf{o3-mini} & 
    \textbf{Qwen3-32B} & 
    \textbf{Baichuan-M1-14B} & %
    \textbf{QuarkMed (Ours)} \\
    \midrule
    MedQA(US)       & \textbf{92.61\%} & 89.23\% & 90.02\% & 88.17\% & 84.02\% & 87.43\% & 74.46\% & \textbf{87.19\%} & 69.83\% & 86.02\% \\
    MedMCQA         & \textbf{82.73\%} & 76.90\% & 79.87\% & 78.01\% & 72.89\% & 76.79\% & 60.57\% & 71.31\% & 65.90\% & \textbf{75.50\%} \\
    PubMedQA        & 76.40\% & 71.80\% & 73.20\% & \textbf{77.25\%} & 75.20\% & 74.80\% & 73.60\% & 73.40\% & 70.80\% & \textbf{79.00\%} \\
    CMExam          & 86.37\% & 78.88\% & 87.50\% & 88.79\% & \textbf{90.10\%} & 86.90\% & 70.74\% & 85.80\% & 76.70\% & \textbf{88.60\%} \\
    AfriMed-QA      & 85.57\% & 82.50\% & \textbf{85.70\%} & 76.57\% & 76.64\% & 81.80\% & \textbf{80.60\%} & 75.90\% & 67.20\% & 74.40\% \\
    MedXpertQA      & \textbf{46.42\%} & 25.90\% & 39.30\% & 30.65\% & 32.63\% & 33.37\% & \textbf{35.43\%} & 26.14\% & 23.00\% & 28.68\% \\
    DiagonissArena  & \textbf{65.91\%} & 51.90\% & 60.65\% & 57.06\% & 54.38\% & 50.00\% & 56.00\% & 52.17\% & 51.50\% & \textbf{61.90\%} \\
    RareBench       & 55.86\% & 55.97\% & 57.56\% & 50.25\% & \textbf{57.98\%} & 49.24\% & 55.06\% & 50.51\% & \textbf{56.46\%} & 52.90\% \\
    MedBullets      & \textbf{82.24\%} & 76.30\% & 82.06\% & 75.53\% & 80.56\% & 78.66\% & \textbf{83.66\%} & 74.14\% & 60.55\% & 77.27\% \\
    CMB-clin        & 3.52    & 3.17    & 3.56    & \textbf{3.70}    & 3.57    & 3.50    & \textbf{3.60}    & 3.50    & 3.07    & 3.50    \\
    MediQ           & \textbf{95.22\%} & 86.70\% & 92.66\% & 84.66\% & 82.81\% & 92.55\% & \textbf{90.78\%} & 87.17\% & 72.40\% & 85.06\% \\
    RedisQA         & \textbf{88.54\%} & 82.30\% & 88.47\% & 85.71\% & 85.99\% & 85.81\% & \textbf{87.35\%} & 82.10\% & 77.10\% & 83.20\% \\
    MMLU(Med)       & \textbf{90.18\%} & 88.20\% & 89.23\% & 88.64\% & 88.73\% & 88.60\% & 87.01\% & 87.48\% & 81.43\% & \textbf{88.37\%} \\
    MedCalc         & \textbf{32.99\%} & 31.31\% & 29.96\% & 32.04\% & 28.38\% & 25.30\% & 35.78\% & 25.53\% & \textbf{38.41\%} & 30.61\% \\
    \midrule
    \textbf{Average}\textsuperscript{*} & \textbf{\textbf{76.36\%}} & {69.80\%} & {74.66\%} & {71.85\%} & {71.40\%} & {71.34\%} & {70.07\%} & {69.02\%} & {63.43\%} & \textbf{\textbf{71.36\%}} \\
    \bottomrule
  \end{tabular}%
  }
 \caption*{\footnotesize\textsuperscript{*} The 4-point scores from the CMB-clin dataset were normalized to a 0-1 range to calculate the overall mean.}
\end{table*}

\begin{table}[ht]
\centering
\caption{Performance on Exam Benchmarks (CPQExam).}
\label{tab:internal_bench_table}
\resizebox{\textwidth}{!}{
\begin{tabular}{llp{1.5cm}p{1.5cm}p{1.5cm}p{1.5cm}p{1.5cm}p{1.5cm}p{1.5cm}p{1.5cm}p{1.5cm}p{1.5cm}} 
\toprule
\textbf{Classification Method} & \textbf{Category} & \textbf{Gemini-2.5-pro-0617} & \textbf{o3-mini} & \textbf{gpt-4o-2024-08-06} & \textbf{DeepSeek-R1-0528} & \textbf{Kimi-k2} & \textbf{Qwen3-235B-A22B-Instruct-2507} & \textbf{Qwen3-235B-A22B} & \textbf{Qwen3-32B} & \textbf{Baichuan-M1-14B} & \textbf{QuarkMed (Ours)} \\
\midrule
Subject & Primary Level & 80.50\% & 70.50\% & 71.67\% & 81.42\% & 83.16\% & 83.03\% & 78.76\% & 78.76\% & 75.67\% & 81.50\% (\textbf{83.3}\%\textsuperscript{*}) \\
& Intermediate Professional & 79.87\% & 66.99\% & 64.15\% & 82.58\% & 74.71\% & 76.79\% & 72.42\% & 70.53\% & 68.04\% & 75.08\% (\textbf{85.4}\%\textsuperscript{*}) \\
& Associate Senior Professional & 65.29\% & 54.25\% & 43.92\% & 72.33\% & 55.42\% & 59.11\% & 56.16\% & 56.15\% & 50.33\% & 66.67\% (\textbf{75.3}\%\textsuperscript{*}) \\
& Senior Professional & 33.60\% & 35.50\% & 12.40\% & 38.70\% & 21.48\% & 27.45\% & 30.34\% & 35.18\% & 30.30\% & 51.70\% (\textbf{67.7}\%\textsuperscript{*}) \\
Question type & Multiple-Choice & 87.45\% & 74.74\% & 75.58\% & 88.14\% & 87.30\% & 87.80\% & 82.08\% & 79.74\% & 78.32\% & 82.58\% (\textbf{91.80}\%\textsuperscript{*}) \\
& Multiple-Response & 27.19\% & 37.15\% & 3.63\% & 32.89\% & 10.60\% & 17.21\% & 26.17\% & 33.90\% & 29.34\% & 55.72\% (\textbf{76.40}\%\textsuperscript{*}) \\
& Shared Stem & 80.06\% & 69.58\% & 63.86\% & 85.31\% & 76.06\% & 76.38\% & 71.34\% & 70.29\% & 59.64\% & 74.10\% (\textbf{81.30}\%\textsuperscript{*}) \\
& Case Analysis & 43.05\% & 36.82\% & 23.74\% & 48.75\% & 32.78\% & 39.47\% & 37.69\% & 38.64\% & 32.21\% & 49.85\% (\textbf{58.50}\%\textsuperscript{*}) \\
\bottomrule
\end{tabular}
}
\caption*{\footnotesize\textsuperscript{*} Knowledge augmentation was employed during the prediction phase.}
\end{table}

\subsection{Results}

As shown in \cref{tab:open_bench_table}, the QuarkMed model demonstrates superior overall performance on medical benchmarks compared to Qwen3-32B \cite{yang2025qwen3technicalreport}, establishing it as one of the most powerful models in its size class.
Notably, the QuarkMed model demonstrates exceptional performance on CPQExam, significantly surpassing other powerful models such as DeepSeek-R1-0528 \cite{deepseekai2025deepseekr1incentivizingreasoningcapability}, o3-Mini, and Gemini-2.5-pro-0617 \cite{comanici2025gemini25pushingfrontier} as shown in \cref{tab:internal_bench_table}. This result underscores the effectiveness of our Reinforcement Learning (RL) training tailored for medical scenarios and highlights the importance of domain adaptation.
Compared to larger open-source models like Qwen-235B-A22B and Kimi-k2, QuarkMed also achieves superior performance on several reasoning datasets, such as MedXpertQA and DiagnosisArena, lagging only behind some prominent closed-source models such as Gemini-2.5-pro-0617. This indicates that our multi-stage training approach, grounded in medical domain knowledge, effectively enhances the model's performance on medical reasoning tasks.

\section{Discussion}
\label{sec:discussion}

This section summarizes practical lessons, current limitations, and forward-looking directions for QuarkMed.

\subsection{Enhancing Performance with Retrieval-Augmented Generation (RAG)}
Although substantial effort was devoted to enriching the model’s parametric (internal) medical knowledge, the strongest and most reliable performance in day-to-day medical assistance and exam-style question answering still hinges on RAG. For high-stakes factual, guideline-timed, or emerging-topic queries, parametric recall alone plateaus: subtle distinctions (e.g., edition-specific dosing updates, recency-dependent public health advisories, or differential refinements) benefit disproportionately from grounding in curated, authority-ranked external sources. In production usage we observe:
\begin{itemize}
  \item Marked gains in factual precision and reduction of subtle hallucinations (e.g., obsolete regimen recommendations) when RAG is enabled, especially for infrequent entities and recently updated contraindications.
  \item Improved calibration: the model is more likely to qualify uncertainty or surface alternative hypotheses when contrasting retrieved passages diverge.
  \item Better exam robustness: for multi-step clinical vignettes, retrieved snippets often disambiguate near-miss distractors (pathophysiology nuance, epidemiologic prevalence shifts) that parametric memory alone confuses.
\end{itemize}
Thus, RAG acts not as an auxiliary enhancement but as a primary reliability layer, and investment priority (index freshness, authority scoring, redundancy pruning, and noise-resilient prompt packaging) remains critical to sustained performance.

\subsection{Implications and Limitations of Reinforcement Learning}
Reinforcement learning (RL) substantially improved structured reasoning (diagnosis selection, test ordering, medication rationality) when clear, automatable verifiers or semi-structured labels existed. Advantages observed:
\begin{itemize}
  \item Reward shaping with hybrid rule+model verifiers amplified format fidelity and reduced reward gaming versus purely model-based preference signals.
  \item Group-wise normalization (GRPO) stabilized multi-task optimization under heterogeneous reward scales.
  \item Curriculum-style difficulty resurfacing prevented early convergence on shallow heuristics.
\end{itemize}
However, limitations persist:
\begin{itemize}
  \item Verifiability Bias: Performance gains concentrate in domains with discrete, checkable endpoints (ICD codes, structured options). Nuanced counseling, longitudinal management planning, lifestyle tailoring, and patient education—where correctness is gradient, contextual, or preference-dependent—remain under-optimized.
  \item Reward Coverage Gaps: Current verifiers incompletely model temporal reasoning (trajectory forecasting, de-escalation strategies), causal justification coherence, and uncertainty articulation.
  \item Overfitting Risk: Tight coupling to deterministic format/verifier schemas risks brittle behavior when schema shifts (new coding versions, guideline reframing).
  \item Sparse / Delayed Feedback: Multi-turn dialogue quality, safety under adversarial probing, and cumulative patient-centric utility lack dense, reliable automatic signals.
  \item Alignment Trade-offs: Maximizing verifiable reasoning occasionally reduces stylistic empathy or brevity unless explicitly multi-objective tuned.
\end{itemize}
Future RL extensions need: (i) semi-verifiable composite rewards (fusing probabilistic factuality estimators with discourse/causal coherence scoring), (ii) active uncertainty elicitation (rewarding calibrated deferral or source citation), (iii) hierarchical RL separating strategic clinical framing from tactical entity selection, (iv) integration of simulation or synthetic patient state transitions for temporal credit assignment, and (v) continual verifier refresh pipelines aligned with evolving guidelines.

\subsection{Future Directions}
\label{subsec:future_work}
Despite its strong performance, the development of QuarkMed highlighted several challenges that point toward future research directions. A key challenge remains the dynamic and ever-evolving nature of medical knowledge. While our RAG system helps, ensuring real-time updates and resolving conflicts between different sources is an ongoing effort. Another limitation is the model's current focus on text-based data. Future work will concentrate on developing multi-modal capabilities, enabling QuarkMed to interpret medical images such as X-rays and pathology slides, which are crucial for many diagnostic workflows.

Furthermore, we aim to improve real-time personalization. Tailoring medical information to an individual's specific health context, while strictly preserving privacy, could significantly enhance the utility of our AI assistant. Finally, we will continue to refine our verification and citation mechanisms. Improving the granularity of citations and developing more robust methods for the model to self-correct and express uncertainty will be critical for building a safer and more reliable medical foundation model.

\section{Conclusion}
\label{sec:conclusion}

This report introduces QuarkMed, a 32-billion parameter foundation model specifically designed for the medical domain. We have detailed our comprehensive, multi-stage approach, which begins with curating a massive and diverse corpus of high-quality medical data. The training methodology combines Supervised Fine-Tuning with two distinct stages of Reinforcement Learning: one focused on verifiable, reasoning-intensive tasks and another on general alignment with human preferences for safety and helpfulness.

The effectiveness of this approach is demonstrated by QuarkMed's state-of-the-art performance on both public and internal benchmarks, including a 70\% accuracy on the Chinese Medical Licensing Examination. By integrating an advanced Retrieval-Augmented Generation system, QuarkMed ensures its responses are grounded in timely and authoritative medical knowledge. As a powerful and versatile AI solution already serving millions of users, QuarkMed represents a significant step forward in developing reliable and effective AI tools for healthcare. We believe this work contributes to the broader goal of leveraging artificial intelligence to improve access to medical information and support better health outcomes globally.

\bibliography{reference}

\end{document}